\documentclass[letterpaper, 10 pt, conference]{ieeeconf}
\IEEEoverridecommandlockouts

\usepackage{cite}
\usepackage{url}

\usepackage{color}

   \usepackage[pdftex]{graphicx}
  \graphicspath{{images/}{pdf/}}
   \DeclareGraphicsExtensions{.pdf,.jpeg,.png}

\usepackage[font=scriptsize]{caption}

\usepackage{multirow}
\usepackage{footnote}

\usepackage{todonotes}

\usepackage{etoolbox}
\makeatletter
\patchcmd{\@makecaption}
  {\scshape}
  {}
  {}
  {}
\makeatother

\begin{document}

%
\title{\LARGE \bf Can we unify monocular detectors for autonomous driving \\by using the pixel-wise semantic segmentation of CNNs?}

\author{Eduardo Romera, Luis M. Bergasa, Roberto Arroyo
 \thanks{*This work was supported by the University of Alcal\'a (UAH) through a FPI grant, the Spanish MINECO through the project SmartElderlyCar (TRA2015-70501-C2-1-R) and the Community of Madrid through the project RoboCity2030
  III-CM project (S2013/MIT-2748). }
  \thanks{The authors are with the Department of Electronics, UAH. Alcal\'a de Henares, Madrid, Spain.
  e-mail: \tt {\small \{eduardo.romera, bergasa, roberto.arroyo\}@depeca.uah.es} 
  }
}



\maketitle

\begin{abstract}

Autonomous driving is a challenging topic that requires complex solutions in perception tasks such as recognition of road, lanes, traffic signs or lights, vehicles and pedestrians. Through years of research, computer vision has grown capable of tackling these tasks with monocular detectors that can provide remarkable detection rates with relatively low processing times. However, the recent appearance of Convolutional Neural Networks (CNNs) has revolutionized the computer vision field and has made possible approaches to perform full pixel-wise semantic segmentation in times close to real time (even on hardware that can be carried on a vehicle). In this paper, we propose to use full image segmentation as an approach to simplify and unify most of the detection tasks required in the perception module of an autonomous vehicle, analyzing major concerns such as computation time and detection performance.


\end{abstract}


%
\IEEEpeerreviewmaketitle

\section{Introduction}

Autonomous driving is a topic with enormous challenges that has recently gathered the interest of teams worlwide in the vehicle industry and research community. Perception is one of its most challenging aspects and requires fusion from multiple sensors to guarantee security by means of redundancies. However, the cheap price and portability of the cameras, the recent advances in computer vision, and the wide possibilities of a high-dimensional signal such as the image (which has not been fully exploited yet) have encouraged the research community to put more efforts on vision-related solutions.
One recent example is Mercedes Bertha \cite{ziegler2014making}, which supposed one of the first autonomous demonstrations on both interurban and urban environments. Researchers managed to avoid the use of LiDAR by developing a complex perception module that highly relies on a stereo camera, RADAR and multiple monocular detectors that are used to cover the deficits of stereo in the tasks of pedestrian, vehicle, traffic light and lane markings detection. 

\begin{figure}[t]
\centering
\includegraphics[width=0.49\textwidth, trim=0cm 0cm 0cm 0cm, clip=true]{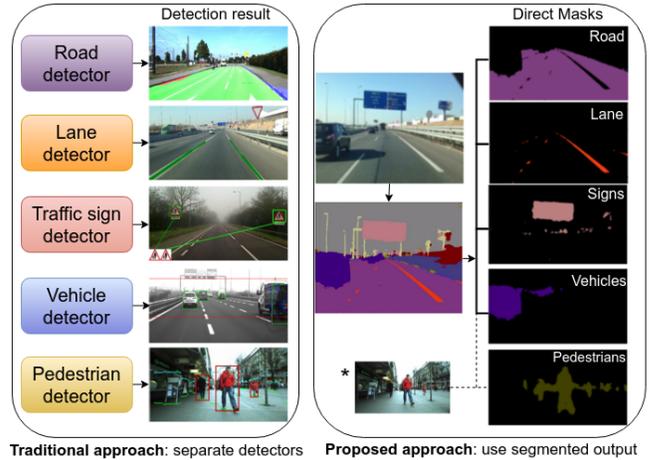}
\vspace{-15pt}
\caption{Two approaches of perception module for autonomous vehicles. 
*A different example image was added for pedestrians (due to lack in the first input image) but these are also segmented by the same system.
}
\vspace{-17pt}
\label{fig:diagram}
\end{figure}

Traditional approaches for monocular object detection were normally based on classification (e.g. SVM, Adaboost) on top of handcrafted features (e.g. Haar, LBP, SIFT, HOG). 
The recent appearance of Convolutional Neural Networks (CNNs) has revolutionized the computer vision field due to the impressive learning capabilities that have been demonstrated in complex visual tasks like recognition. The work in \cite{girshick2014rich} presented an approach to combine regions with CNN features (R-CNN) for object detection, that overpassed in more than 30\% previous best results in PASCAL VOC \cite{everingham2010pascal}. 
More recently, \cite{long2015fully} proposed to adapt state-of-the-art CNNs by converting them to Fully Convolutional Networks (FCN) that can be trained end-to-end to perform pixel-wise semantic segmentation. However, an issue with this adaptation is the large size of the receptive fields in the convolutional filters of these networks, which results in coarse outputs with poorly delineated objects. Works like \cite{chen2014semantic} and \cite{zheng2015conditional} proposed to solve these problems by combining the network with Conditional Random Fields. While these approaches produce remarkable segmentation, their processing times are a bit far from real-time by ranging from 0.5 to 4 seconds per image on high-end GPUs. However, very recent works like \cite{badrinarayanan2015segnet} (SegNet) have proposed to combine convolution and deconvolution layers in a new architecture that is optimal for road scene understanding, as it excels in modeling appearance, shape and spatial-relationships (context), producing similar segmentation accuracy with much lower processing times. 

These advances have made possible the use of full scene segmentation in time-critical applications like autonomous driving. In this paper, we study how to cover the tasks of the perception module of a vehicle by unifying detection tasks by means of the output of a pixel-wise semantic segmentation network (see Fig. \ref{fig:diagram}).
The main difference of such approach is that, instead of performing classification on pre-selected region candidates (traditional methods), the whole image is classified to use this output as base to higher-level understanding algorithms required for vehicle navigation (e.g. collision avoidance). In the following sections we analyze the advantages and challenges of the proposed approach as compared to traditional monocular detection approaches.

\section{Discussion on general concerns}\label{sec:2}

We analyze the general advantages and disadvantages of unifying detectors by using pixel-wise segmentation:

\subsubsection{\textbf{Computational cost and latency}} Until recently, full pixel segmentation was not usable in terms of efficiency. However, recent works such as SegNet \cite{badrinarayanan2015segnet} allow to compute decent segmentation (see Fig.~\ref{segnet2}) in less than 100ms on a single GPU. If the task is only to detect one single category, then any monocular detector can outperform this performance. However, in autonomous vehicles, it is presumed that we are interested in detecting everything that surrounds the vehicle. In this case, running all detector jointly scales computation much further than 100ms (see Table~\ref{table1}), so the unification in this case is advantageous. A critical point may be the latency: fast moving objects are response-time critical (e.g. pedestrians in street scenarios), and the unification detriments these cases as the latency is the same for all detections.

\subsubsection{\textbf{Implementation complexity}} One of the main advantages of the unification is that it simplifies the problem. In the case of independent detectors, each approach is based on a different model, algorithm and training, which implies high complexity for implementation and for developing algorithms on top of this result. Additionally, optimization is not always straightforward (e.g. parallelization or GPU implementation). However, in the case of using the segmented output of a CNN, 
every implementation challenge unifies within the same module.
Recent deep learning libraries like Caffe \cite{jia2014caffe} have simplified development (shared code), optimization (direct GPU implementation) and even have reinforced the community implication on sharing pre-trained models (Model Zoo). Upgrading the system is also straightforward, both in the case of changing network architecture and in the case of re-training with new data (easy to fine-tune parameters instead of having to re-train multiple classifiers). 



\subsubsection{\textbf{Output quality}} In the case of pixel-wise segmentation, even if the output is not precise, the result might still be of use. For instance, in obstacle detection it is preferred to know that there is a pedestrian in some part of the image (even if the shape is not accurate) to not having anything detected at all. Additionally, for cases in which shape is important (i.e. traffic signs) the segmented output may be used as a "ROI detector" to scan object candidates and apply a simple classifier in that area (with reduced computational cost). Moreover, having the full pixel segmentation provides additional advantages which are not only related to detection, but also to higher-level applications such as semantic reconstruction of the environment.

\begin{figure}[t]
	\centering
	\includegraphics[width=0.48\textwidth, trim=0cm 0cm 0cm 0cm, clip=true]{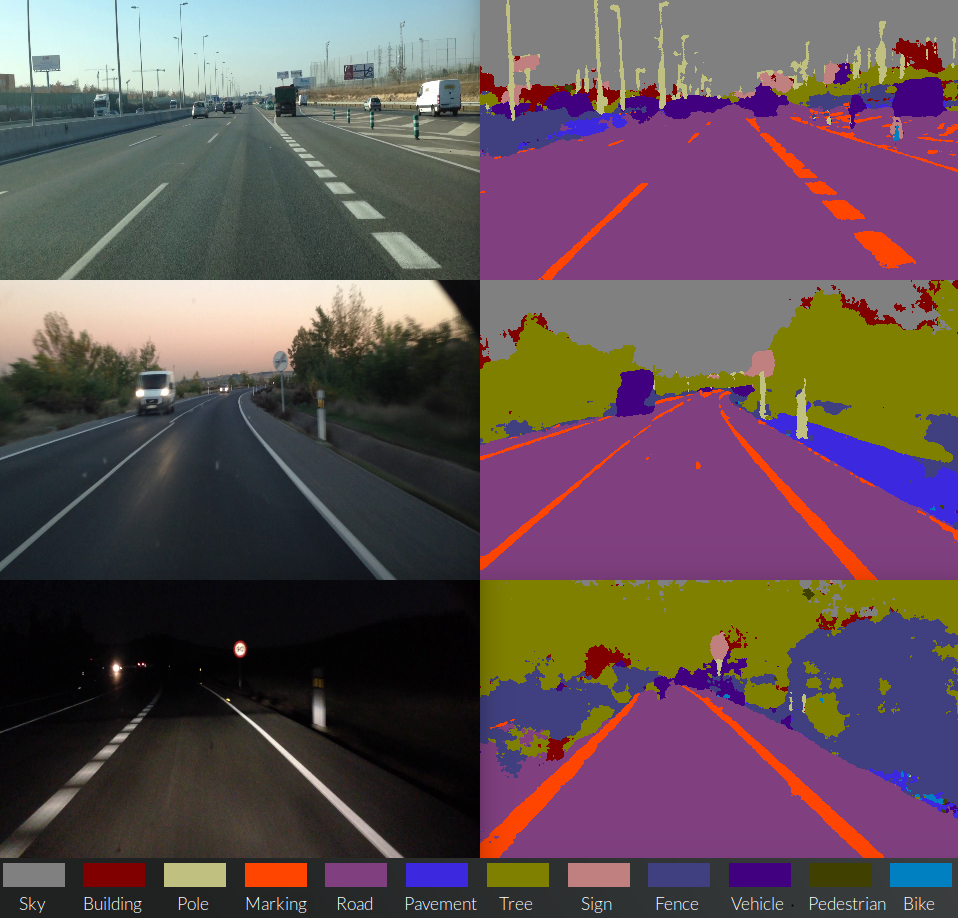}
	\caption{Example cases of pixel-wise segmentation performed by SegNet on real road scenarios: daytime (top), dusk (middle) and nighttime (bottom)}
	\vspace{-15pt}
	\label{segnet2}
\end{figure}

%

\subsubsection{\textbf{Data availability and training}} 
In the case of traditional detectors, each one requires a set of positives and negatives to train each specific class. With the recent progress on CNNs, pixel-annotated datasets are becoming popular and more detailed, so training end-to-end to perform pixel segmentation has become feasible and a good practice to jointly optimize the network weights. PASCAL VOC 2012~\cite{everingham2010pascal} is one of the first with fully labeled images (indoor and outdoor). More recently, datasets oriented towards vehicle environments have emerged, like CamVID~\cite{brostow2009semantic} 
and Cityscapes~\cite{Cordts2016Cityscapes}. 


\subsubsection{\textbf{Tracking}} 
While traditional approaches like \cite{girshick2014rich} firstly extract region candidates for posterior classification, 
in the proposed approach no region candidates are extracted before classification (segmentation) occurs. Therefore, if tracking of individual objects is desired, then the object candidates must be extracted a posteriori. Some possibilities are to extract object candidates by applying shape constraints or enclosing those with same motion (e.g. using optical flow).
Additionally, the unified knowledge about the whole road environment allows for better application of inter-class relations to improve tracking.
For instance, some vehicle detectors like \cite{romera2015} rely on lane detection to track according to geometrical constraints and to filter impossible vehicle locations. By having full segmentation, these relations are easier to apply (see Fig.~\ref {segnet3} for an example). In fact, the spatial-relationship between different classes, such as the difference between road and side-walk, is implicitly learned by the CNN.

%

\subsubsection{\textbf{Occlusions}} Works like \cite{dollar2012pedestrian} reflect that occlusions suppose a big impact on detection performance on traditional detectors, even for partial occlusions, and it surprisingly affects similarly to both holistic and part-based detectors. By using the pixel-wise segmented output this effect is reduced, as we do not need region candidates that enclose the whole object, but even a small object region can produce enough feature activations to segment correctly the class on those pixels (unless the occlusion occurs in the parts with the most representative features of the specific class). 

\subsubsection{\textbf{Challenging lighting conditions}} While detection decay is severe in any traditional detector at challenging conditions (night or adverse weather), the accuracy in the segmentation produced by a CNN remains still reasonable (see Fig.~\ref{segnet2} for dusk and nighttime examples). At night, areas covered by the vehicle headlights (road and close signs) are correct, while distant objects might be extremely noisy and may require combination with additional detection methods.

\section{Discussion on specific detectors}\label{sec:3}

\begin{figure}[t]
	\includegraphics[width=0.49\textwidth, trim=0cm 0cm 0cm 0cm, clip=true]{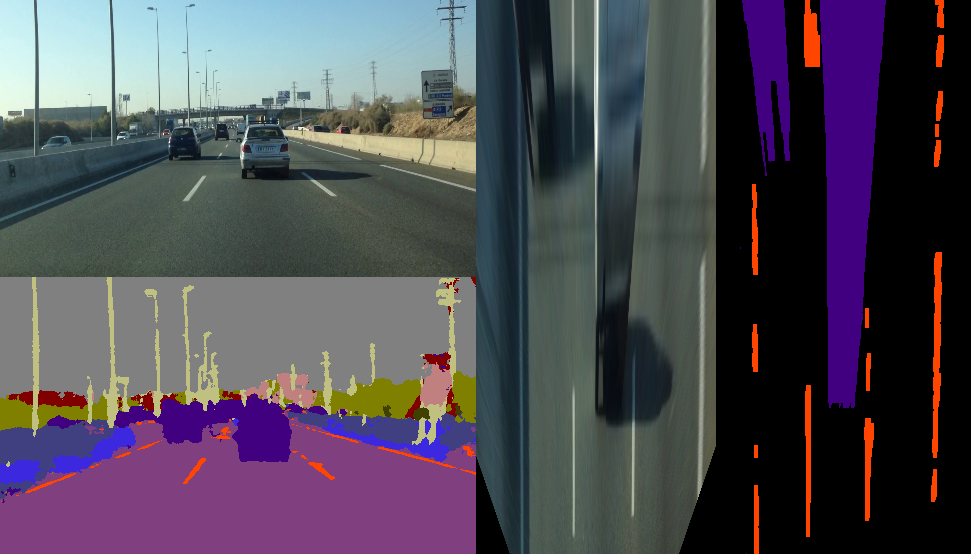}
	\caption{Example application of the segmented output (direct extraction of lane and vehicle positions). Left: input and segmented output. Mid: bird's view (IPM) of input. Right: bird's view of segmented lanes and vehicles.}
	\vspace{-15pt}
	\label{segnet3}
\end{figure}

\begin{table*}[]
\centering
\caption{
Summarized quantitative comparison between monocular detectors (left) and pixel-wise segmentation networks (right). It should be considered only as an overview, as direct performance comparison is not possible due to different evaluation metrics on each task. *Bergasa\cite{bergasa2014drivesafe} and Romera\cite{romera2015} are smartphone approaches, but times are measured in computer by using an iPhone 6 simulator. **IoU means Intersect over Union = TP/(TP+FP+FN), metric introduced in \cite{everingham2010pascal} to measure pixel classification accuracy.}
\label{table1}
\resizebox{\textwidth}{!}{
\begin{tabular}{|c|c|c|c|l|c|c|c|}
\cline{1-4} \cline{6-8}
\textbf{Task}                         & \textbf{Method}         & \textbf{Time per frame {[}environment, resolution{]}}   & \textbf{Performance (different metrics)}                                                                                   &  & \textbf{Network}                       & \begin{tabular}[c]{@{}c@{}}\textbf{Time per frame} \\ (forward pass on 500x500)\end{tabular}                          & \begin{tabular}[c]{@{}c@{}}\textbf{Performance IoU**}\\ (Top: PASCAL VOC \cite{everingham2010pascal}, \\ Bot: Cityscapes \cite{Cordts2016Cityscapes})\end{tabular} \\ \cline{1-4} \cline{6-8} 
\textbf{Road}                         & Passani \cite{passani2015fast}                 & 50 ms {[}4 cores i7@2.1Ghz, 375x1242{]}                 & \begin{tabular}[c]{@{}c@{}}Precision: 78-89\%\\ Recall: 82-93\%\end{tabular}                                               &  & \multirow{2}{*}{\textbf{FCN-8s} \cite{long2015fully}}   & \multirow{2}{*}{500 ms (published time)}                                                                               & 62.2\% class avg                                                                                       \\ \cline{1-4} \cline{8-8} 
\multirow{2}{*}{\textbf{Lane}}        & Bergasa \cite{bergasa2014drivesafe}               & 5 ms {[}4 cores i7@2.2Ghz*, 640x480{]}                  & n/a                                                                                                                        &  &                                    &                                                                                                                        & \begin{tabular}[c]{@{}c@{}}65.3\% class avg\\ 85.7\% category avg\end{tabular}                         \\ \cline{2-4} \cline{6-8} 
                                      & Kim  \cite{kim2008robust}                   & 25 ms {[}Intel Core@1.83Ghz, 70x250{]}                  & n/a                                                                                                                        &  & \multirow{2}{*}{\textbf{Deeplab} \cite{chen2014semantic}}  & \multirow{2}{*}{4000 ms (published time)}                                                                              & 58\% class avg                                                                                         \\ \cline{1-4} \cline{8-8} 
\textbf{Signs}                        & (various) \cite{mogelmose2012vision}              & N/A                                                     & \begin{tabular}[c]{@{}c@{}}72 to 99\% mean \\ detection rate\end{tabular}                                                  &  &                                    &                                                                                                                        & \begin{tabular}[c]{@{}c@{}}64.8\% class avg\\ 81.3\% category avg\end{tabular}                         \\ \cline{1-4} \cline{6-8} 
\multirow{2}{*}{\textbf{Vehicles}}    & Romera  \cite{romera2015}                & 16 ms {[}4 cores i7@2.2Ghz*, 640x480{]}                 & \begin{tabular}[c]{@{}c@{}}Prec: +90\%\\ Recall (\textless60m): 80 to 95\%\end{tabular}                                    &  & \multirow{2}{*}{\begin{tabular}[c]{@{}c@{}}\textbf{CRFasRNN}\\ \cite{zheng2015conditional}\end{tabular} } & \multirow{2}{*}{\begin{tabular}[c]{@{}c@{}}700 ms (published time)\\ $\sim$2 sec. (tested Nvidia Titan X)\end{tabular}}      & 69.6\% class avg                                                                                       \\ \cline{2-4} \cline{8-8} 
                                      & Satzkoda \cite{satzodamultipart}          & 50 ms (4 cores i7, 1024x768{]}                          & \begin{tabular}[c]{@{}c@{}}Prec: 94\%\\ Recall: 97\%\end{tabular}                                                          &  &                                    &                                                                                                                        & \begin{tabular}[c]{@{}c@{}}62.5\% class avg\\ 82.7\% category avg\end{tabular}                         \\ \cline{1-4} \cline{6-8} 
\multirow{2}{*}{\textbf{Pedestrians}} & \multirow{2}{*}{Dollar \cite{dollar2012pedestrian}} & \multirow{2}{*}{150-350 ms {[}2012 CPU n/a, 640x480{]}} & \multirow{2}{*}{\begin{tabular}[c]{@{}c@{}}57\% log-average \\ miss rate (on clearly \\ visible pedestrians)\end{tabular}} &  & \multirow{2}{*}{\textbf{SegNet} \cite{badrinarayanan2015segnet}}   & \multirow{2}{*}{\begin{tabular}[c]{@{}c@{}}60 ms (published time)\\ $\sim$100 ms (tested Nvidia Titan X)\end{tabular}} & 59.1\% class avg                                                                                       \\ \cline{8-8} 
                                      &                         &                                                         &                                                                                                                            &  &                                    &                                                                                                                        & \begin{tabular}[c]{@{}c@{}}57.0\% class avg\\ 79.1\% category avg\end{tabular}                         \\ \cline{1-4} \cline{6-8} 
\end{tabular}
}
\vspace{-5pt}
\end{table*}

\begin{table*}[]
\centering

\caption{\vspace{-3pt}\scriptsize Pixel accuracy per class (\%) of SegNet trained on a multiple dataset (3.5K images) evaluated on CamVid \cite{brostow2009semantic} dataset (11 classes)}
\label{table2}
\footnotesize
\resizebox{!}{10pt}{
\begin{tabular}{|c||c|c|c|c|c|c|c|c|c|c|c||c|c|c|}
\hline
\multirow{2}{*}{\begin{tabular}[c]{@{}c@{}}SegNet\\ (3.5K)\end{tabular}} & Build. & Tree & Sky  & Car  & Sign & Road & Pedes. & Fence & Pole & Sidew. & Bike & \begin{tabular}[c]{@{}c@{}}Class Avg.\end{tabular} & \begin{tabular}[c]{@{}c@{}}Global Avg.\end{tabular} & \begin{tabular}[c]{@{}c@{}}Mean I/U\end{tabular} \\
                                                                         & 73.9  & 90.6 & 90.1 & 86.4 & 69.8 & 94.5 & 86.8   & 67.9  & 74.0 & 94.7   & 52.9 & 80.1                                                 & 86.7                                                  & 60.4                                               \\ \hline
\end{tabular}
}
\vspace{-10pt}
\end{table*}

We discuss the advantages and drawbacks on specific detection concerns in five of the main categories an autonomous system should perceive. Table~\ref{table1} presents a quantitative comparison that summarizes the reviewed methods.
\setcounter{subsubsection}{0}
\subsubsection{\textbf{Road}}
Detecting the road is necessary for delimiting the driving space. Bertha \cite{ziegler2014making} covered this part with stereo vision by treating ground plane as the road. However, stereo vision has its drawbacks, as its range may not be sufficient for high speeds (i.e. motorways) and the ground plane may not necessarily be asphalt road in some environments (e.g. grass, low sidewalks or water). Some works have treated road detection with monocular cameras like \cite{passani2015fast}, that achieved near real-time pixel-wise output: around 50ms per frame (4 cores@2.8GHz) with a precisions between 77\%-88\% and recall rates between 83\%-96\%. However, the computation impact is still high considering that all the cores of a high-end CPU are being used to only detect the road. Therefore, in this case there is no specific reason for using an independent detector for road instead of the segmented output of a CNN, which in fact provides better detection performance (94.5\% pixel accuracy in the road class, as seen in Table~\ref{table2}).


\subsubsection{\textbf{Lane markings}}
Detecting own lane is required for avoiding invasion of other lanes which may be occupied by other participants or that may be reserved for the opposite direction. Current vehicles like \cite{ziegler2014making} rely the knowledge of lane spatial properties to a pre-designed map. There has been active research on lane detection algorithms with a wide variety of lane representations (e.g. fixed-width line pairs, spline ribbon), detection and tracking techniques (from Hough transform to probabilistic fitting and Kalman filtering). 
A simple approach for smartphones that used a clothoidal model on a EKF (based on points extracted from an adaptative Canny algorithm) is presented in \cite{bergasa2014drivesafe}. A robust probabilistic approach suited for all kind of challenging scenarios (worn lanes and different kinds of splitting lanes) was presented in \cite{kim2008robust}. Computation cost in these detectors is low. \cite{bergasa2014drivesafe} can run at 30 FPS on a single core on iPhone~5, and \cite{kim2008robust} in less than 25ms per frame on an Intel Core@1.83GHz (2008), even faster in recent processors. 
Because of this, using full segmentation only to extract lanes could be disproportionate,
but it makes sense if the segmentation is used for several tasks as in the proposed approach. From the segmented output, it is straightforward to obtain a lane model by using an EKF (with clothoidal model for curvature) on the segmented lane-markings observations from a bird's view perspective (like in Fig.~\ref{segnet3}). An additional concern are the roads with no visible markings. Bertha used a curb detector to consider them a boundary line (mainly for suburban areas). In the case of full segmentation, approaches like SegNet allow to separate classes like sidewalk and fence, so in combination with the road detection it would be simple to detect them as a lane boundary if there is no lane marking.


\subsubsection{\textbf{Traffic signs and lights}}
Detecting traffic signs and lights is needed for understanding the main navigation rules. However, this is a challenging task due to the high sign variability, different regulations and the spatial implications (e.g. which signs apply to which lanes). Reader may refer to \cite{mogelmose2012vision} for a study on traffic sign recognition methods. Current autonomous vehicles rely all this information to highly detailed maps and they only detect dynamic properties like the states of traffic lights. 
The main problem of an approach based on full segmentation is that no sign type classification is performed, so the segmentation could help in locating possible sign/light candidates in the image, but still an additional algorithm would be required to extract information about the properties. Therefore, there is no direct computation saving in the classification on the proposed approach (only in the extraction of sign candidates). However, applying a simple classification to infer the sign type or the state of a traffic light is relatively simple if the region candidate is provided.


\subsubsection{\textbf{Vehicles}}
It is essential to detect other traffic participants that share the road (i.e. cars, trucks, buses, bicycles and motorbikes) to avoid collision and also to understand their position and intentions in order to anticipate and respect their priority (e.g. not invade their driving lane). Recently, \cite{romera2015} introduced a detection and tracking framework for smartphones by applying geometrical constraints and using multiple regions to reduce scanned amount of pixels. However, this results in strong decay in detection from 70 meters onwards, which may not be usable for autonomous driving. Other approaches like \cite{satzodamultipart} propose a two-part detection technique using an active learning framework and symmetry-based iterative analysis, providing state-of-the-art results in around 20FPS on 4 cores (i7). 
The problem of these approaches is that they are normally trained and tested only on cars (and trucks at most). Different vehicles have varied shapes, which results in an important detection decay even between similar vehicles (see \cite{romera2015} for car vs trucks), and other kinds like motorbikes or bicycles will not be detected at all. In such case, separate classifiers trained for each vehicle category would be desired, but it is not realistic due to computational constraints.
Due to these reasons, the proposed approach based on full segmentation would be advantageous as it differs and detects varied kinds of vehicles (as required by an autonomous vehicle) with remarkable detection rates (see Table~\ref{table2}). Distinguishing between type of vehicles also adds the possibility to use different prediction models (e.g. a car moves differently to a bike). Additionally, the simultaneous knowledge of the road and lanes promotes easier tracking (known vehicle position with respect to lanes, as in Fig.~\ref{segnet3}) or filtering of impossible locations (e.g. vegetation, fence).

\subsubsection{\textbf{Pedestrians}}
Robust pedestrian detection is a key aspect in autonomous driving, specially in urban scenarios. The survey \cite{dollar2012pedestrian} presents an extensive evaluation of state-of-the art pedestrian detectors in a unified framework. Nearly all modern detectors employ as features some form of gradient histograms, with SVM and boosted classifiers as the most popular learning algorithms. A widely known detector, with one of the best detection performances and fastest computation is FPDW~\cite{dollar2012pedestrian}, processed between 2.7 and 6.5 FPS on CPU (640x480).
Although additional improvements can enhance this result by speeding the pre-selection of region candidates or the detection itself, this task still supposes heavy computation load to the framework and the detection rates may not be robust enough for autonomous driving (miss ratio is still significant and there is a heavy detection decay with occlusions and candidates of small pixel sizes). Therefore, relying the pedestrian detection task on the pixel-wise segmentation (proposed approach) would be advantageous in both computation (see Table~\ref{table1}) and detection performance (86.8\% pixel accuracy as seen in Table~\ref{table2}).

\section{Conclusion and Future Work}

This work supposes a prior study to design a perception approach for autonomous vehicles fully based in segmentation performed by CNNs, which will be reviewed in future works. 
The aim is to simplify things in terms of implementation and to solve open challenges of traditional detectors such as the high combined computation cost and detection decay in cases like occlusion. 
However, such an approach also adds challenges of its own (achieved segmentation is not perfect), but end-to-end CNNs are a recent topic and it still has room for future improvements in both efficiency and accuracy, where new datasets and network architectures will play an essential role.




\bibliographystyle{IEEEtran}
\bibliography{IEEEabrv,bib}

\end{document}